\documentclass[runningheads]{llncs}
\usepackage{enumitem}

\usepackage[T1]{fontenc}
\usepackage{url}
\usepackage{booktabs}
\usepackage{graphicx}
\begin{document}
\title{LLM Driven Processes to Foster Explainable AI}
\titlerunning{LLM Driven Processes to Foster Explainable AI}
\author{Marcel Pehlke \and Marc Jansen}
\authorrunning{Pehlke et al.}
\institute{University of Applied Sciences Ruhr West, Bottrop, Germany\\
\email{\{marcel.pehlke, marc.jansen\}@hs-ruhrwest.de}\\
}
\maketitle
\begin{abstract}

We present a modular, explainable LLM-agent pipeline for decision support that externalizes reasoning into auditable artifacts. The system instantiates three frameworks—Vester’s Sensitivity Model (factor set, signed impact matrix, systemic roles, feedback loops), normal-form games (strategies, payoff matrix, equilibria), and sequential games (role-conditioned agents, tree construction, backward induction), with swappable modules at every step. LLM components (default: GPT-5) are paired with deterministic analyzers for equilibria and matrix-based role classification, yielding traceable intermediates rather than opaque outputs. In a real-world logistics case (100 runs), mean factor alignment with a human baseline was \textbf{55.5\%} over 26 factors and \textbf{62.9\%} on the transport-core subset; \textbf{role agreement} over matches was \textbf{57\%}. An LLM judge using an eight-criterion rubric (max 100) scored runs on par with a reconstructed human baseline. Configurable LLM pipelines can thus mimic expert workflows with transparent, inspectable steps.

\end{abstract}

\section{Introduction}
Canada is currently facing one of the most pressing social and economic challenges of the decade: the housing crisis. The federal government recently announced its Build Canada Homes initiative, pledging to accelerate the construction of affordable housing to address shortages across the country \footnote{Government of Canada: ``Prime Minister Carney launches Build Canada Homes,'' September 14, 2025. Available at: \url{https://www.pm.gc.ca/en/news/news-releases/2025/09/14/prime-minister-carney-launches-build-canada-homes}}. When asking today’s most advanced language models about potential solutions, responses are typically broad and multi-factored, pointing to issues ranging from zoning policies to labor shortages. While such answers highlight the multifaceted nature of the problem, they often remain vague, lack prioritization, and most importantly, provide no transparent reasoning path. Users are left with plausible statements, but without an explanation of how the model arrived at them. This “black-box factor” undermines trust in LLM-driven decision support, particularly for domains where systemic reasoning and accountability are essential.

At the same time, the field of artificial intelligence is undergoing a notable paradigm shift. Instead of focusing solely on scaling up models, recent research emphasizes reasoning and orchestration. Reasoning models, showcase stepwise problem solving through chain-of-thought techniques, making intermediate reasoning explicit \cite{10.5555/3600270.3602070}. Orchestration mechanisms further extend this principle by dynamically routing queries to the most suitable process or model, enabling adaptive trade-offs between accuracy, latency, and cost \cite{mohammadshahi2024routoolearningroutelarge}. Building on these trends, LLM-based agents introduce an additional layer: they do not only reason, but also act, decompose tasks, and coordinate across specialized roles. Multi-agent setups in particular promise structured, auditable processes that move beyond opaque one-shot answers \cite{ijcai2024p890}.

This work leverages that paradigm shift by implementing three structured processes within an LLM agent system. Vester’s Sensitivity Analysis, normal-form game analysis, and a sequential (extensive-form) game analysis. Each framework provides a transparent reasoning workflow, externalizing intermediate artifacts such as influence matrices, payoff tables, and decision trees. Our aim is not to render the LLM’s internal computations transparent, but to add a process layer on top of the LLM, in which the process allows detailled statistical analysis providing futher insides into the decision. By embedding the model in structured processes (e.g., Vester matrices, game trees), we expose the macro-level reasoning: what inputs were used, which options were considered, how criteria were applied, and why a path was selected.

From this perspective, we pursue the following research question:
Can an LLM-based agent system mimic established decision-making frameworks such as game theory and sensitivity analysis, thereby providing a traceable reasoning process instead of opaque outputs?

To address this question, we implement and evaluate the proposed agent pipelines on real-world inspired scenario. Our evaluation strategy compares LLM-generated results against human-created baselines and methodological expectations. Specifically, we run the Vester sensitivity analysis pipeline 100 times on a logistics case study and benchmark the outputs against an established human-conducted study. In doing so, we do not only compare the final results of both studies, but also assess the quality of execution, that is, how well the LLM pipeline mimics the structure of the sensitivity analysis. This allows us to determine whether the system can reliably reproduce the process of Vester’s methodology, rather than producing isolated or coincidental overlaps.
\section{State of the Art}
This chapter reviews the state of the art in large language models and structured decision frameworks, providing the foundations for our proposed agent-based approach.

\subsection{Large Language Models: Capabilities, Limits, and Structured Reasoning}

Large Language Models (LLMs) such as GPT-4 and GPT-5 have become central to AI research and application. Their foundation lies in the Transformer architecture \cite{10.5555/3295222.3295349}, which enabled scalable training and parallelization. Scaling laws and model families like GPT, PaLM, and Chinchilla revealed that size and data could yield emergent abilities, but also that performance gains now plateau \cite{DBLP:journals/corr/abs-2001-08361,10.5555/3600270.3602446}. Despite their impressive capabilities, LLMs remain opaque: they frequently hallucinate, producing plausible but incorrect statements, and provide no clear reasoning path, which undermines their explainability \cite{info:doi/10.2196/53164}.

Recent work shifts focus from scaling to structure. Chain-of-Thought (CoT) prompting \cite{10.5555/3600270.3602070} improves accuracy on multi-step problems by exposing intermediate reasoning. OpenAI’s reasoning models, notably o1 and o3-mini, refine this approach by offering adjustable “reasoning effort” levels. The o1 model reached PhD-level proficiency in subjects like physics, chemistry, and biology, solving 83\% of problems on the American Invitational Mathematics Examination (AIME) and ranking in the 89th percentile in Codeforces competitions. Its successor, o3-mini, matches this performance at medium effort while delivering faster responses, and with high effort achieved 83.6\% accuracy on AIME 2024 and 77\% on PhD-level science benchmarks (GPQA Diamond) \footnote{OpenAI: ``Learning to Reason with LLMs,'' September 12, 2024. Available at: \url{https://openai.com/index/learning-to-reason-with-llms/}}.

While CoT enhances single-model reasoning, many tasks require selecting which model to use at runtime. GPT-5 exemplifies this with a modular setup: a fast base model for routine queries, a slower reasoning model for complex tasks, and a router that dynamically directs queries based on complexity and user signals. This orchestration allows efficient trade-offs between accuracy, latency, and cost \footnote{OpenAI: “Introducing GPT-5,” 7. August 2025. Available at: \url{https://openai.com/index/introducing-gpt-5/}}. Conceptually, this is closely aligned with the idea pursued in this work: different prompts can trigger not only different models, but also different structured processes, such as Vester’s sensitivity analysis or game-theoretic reasoning, depending on the nature of the problem. In both cases, orchestration enables dynamic adaptation by selecting the most appropriate reasoning pathway for each task, rather than relying on a single, uniform approach.

Similar ideas appear in research such as the Leeroo Orchestrator, which treats entire LLMs as “experts” and routes queries to the most suitable one \cite{mohammadshahi2024routoolearningroutelarge}. Instead of building even-larger models, orchestration leverages diverse models, turning model selection itself into a learned decision process. This signals a paradigm shift: performance stems not only from scaling individual networks but from coordinating specialized components under adaptive routing. Orchestration thus represents a crucial development for deploying LLMs in real world, resource sensitive environments.

Building on orchestration, LLM agents extend reasoning into acting. They interpret tasks, decompose them into steps, retrieve knowledge, and use tools. Frameworks such as ReAct \cite{ac1f09077393404a8bea5141d8710259} and Tree-of-Thoughts \cite{10.5555/3666122.3666639} highlight how agents combine reasoning with environment interaction in continuous sense–plan–act cycles. Multi-agent setups further show how LLMs can emulate teamwork by assigning roles and cross-validating ideas \cite{ijcai2024p890}. Importantly, modular agent designs externalize intermediate artifacts, plans, decisions, and memory states, that improve explainability and make reasoning auditable.

A key example of embedding explainability into decision-making is the Questions–Options–Criteria (QOC) framework \cite{MacLean01091991}. Recent work demonstrates that QOC can be effectively integrated with AI. Schmidt et al. \cite{10.1007/978-3-031-71739-0_27} propose an AI-augmented QOC process where models contribute at three levels: (1) as assistants that suggest new options or criteria from data, (2) as participants evaluating alternatives alongside humans, or (3) as autonomous decision-makers in multi-agent setups that generate and assess the full QOC structure. Importantly, because each option and assessment remains explicitly logged, the reasoning pathway is auditable, even when agents act autonomously. Their evaluation highlights that this approach improves both traceability and decision quality, aligning with the goals of Explainable AI (XAI).  

This shows how structured frameworks can complement LLMs: they externalize reasoning into human-auditable steps, reduce black-box opacity, and provide a verifiable link between input, reasoning, and outcome. Extending this principle to other methods, such as Vester’s sensitivity analysis or game-theoretic reasoning, illustrates a path toward AI agents that not only generate powerful results but also justify them through transparent, structured processes.

\subsection{Frederic Vester’s Sensitivity Model (VSM)}

Frederic Vester’s Sensitivity Model (VSM) is a systems method for tackling complex, interdependent problems by combining qualitative judgement with lightweight quantification and an explicit focus on feedback and roles \cite{article}. Eight bio-cybernetic principles (e.g., dominant negative feedbacks, diversity/decentralization, avoiding growth dependence) guide a structured workflow that surfaces key drivers, sensitivities, and leverage points \cite{etde_8137136,10.1007/978-3-319-07953-0_3}.

\begin{enumerate}[label=\textbf{\arabic*.}]
\item \textbf{System Description} – define boundaries, goals, and context.
\item \textbf{Variable Set} – identify 10--20 essential variables.
\item \textbf{Criteria Matrix} – filter variables by systemic relevance (e.g., measurability, influenceability).
\item \textbf{Impact Matrix} – score how each variable influences the others.
\item \textbf{Systemic Role Analysis} – classify variables (drivers, critical, buffering, reactive) via active/passive sums.
\item \textbf{Effect System} – visualize the network and feedback loops.
\item \textbf{Partial Scenarios} – explore qualitative “what-if”s on subloops/subsystems.
\item \textbf{Simulation} – run qualitative/fuzzy simulations of scenario dynamics.
\item \textbf{Cybernetic Evaluation} – assess strategies against the principles for viability.
\end{enumerate}

Among these, the \textbf{impact matrix} and \textbf{role analysis} are the methodological core, reducing complexity by revealing drivers, responses, and leverage points \cite{10.1007/978-3-319-07953-0_3}. Embedding this workflow in an LLM multi-agent pipeline yields traceable intermediate artifacts (matrices, roles, loops), enabling problem solving with visible reasoning rather than black-box outputs.

\subsection{Game Theory Aspects}

Game theory studies strategic interactions where outcomes depend on the choices of multiple agents. Formalized by von Neumann and Morgenstern in 1944 \cite{VonNeumann+Morgenstern:1944}, it has become a central tool in economics and beyond. Two core representations are the \emph{normal form} (simultaneous moves) and the \emph{extensive form} (sequential moves).

In a normal-form game, each player chooses a strategy simultaneously and payoffs are defined for every strategy profile. A classical example is the Prisoner’s Dilemma \cite{repec:inm:ormnsc:v:5:y:1958:i:1:p:5-26}, where individually rational choices (defection) lead to a collectively suboptimal outcome. The central solution concept here is the Nash equilibrium \cite{540b73bd-a3f1-333e-a206-c24d0fbbb8bc}: a strategy profile in which no player can improve their payoff by unilaterally deviating. Nash proved that every finite game has at least one equilibrium, possibly in mixed strategies, making it the cornerstone of non-cooperative game theory.

Many interactions are sequential. The extensive form represents games as decision trees where players act in turn. In this setting, Nash equilibrium may allow non-credible strategies. Selten introduced the refinement of subgame perfect equilibrium (SPE) \cite{142260fb-4348-369a-a33c-e562907cef9b}, requiring that strategies form an equilibrium in every subgame. For perfect-information games, backward induction provides a constructive way to compute SPE, ensuring credibility of strategies at every node.

Normal-form and extensive-form reasoning offer structured ways to analyze interdependent decisions. Concepts like Nash equilibrium and backward induction can be operationalized in LLM-based agents, enabling them to “play through” strategic scenarios step by step. Embedding such frameworks allows AI agents not only to produce outcomes, but to expose their reasoning paths through well-established game-theoretic solution concepts—making decisions both analyzable and traceable.

\section{Architecture}
The system is designed in a modular way, where each analytical framework is implemented as a self-contained process. Specifically, this work implements three processes: Vester’s Sensitivity Analysis, normal-form game analysis, and a sequential (extensive-form) game analysis. Each process follows its own structured workflow, from input parsing to stepwise reasoning and final output.   

A future orchestration layer, which is not implemented here, could route each user prompt to the most suitable process, mirroring recent runtime orchestration for LLMs. Because the processes are modular, the orchestrator can scale and evolve, and via MCP it could even dispatch to processes running on different machines while retaining only lightweight location metadata.

In this work, the focus is placed on implementing the individual processes themselves rather than the orchestration layer. Nonetheless, the modular architecture demonstrates how structured analytical methods can be integrated into LLM-based systems, providing both problem-solving capability and transparency in the reasoning process.

\section{Implementation}
Across all three processes (Vester sensitivity analysis, normal-form games, and sequential games), the implementation follows a modular, step-wise design: each step is an independent module with swappable AI agents (model family, temperature, reasoning effort, etc.), allowing fine-grained configuration and future upgrades. At the end of each process, a dedicated summarization agent consolidates the intermediate artifacts into a human-readable explanation and, where applicable, produces visuals (e.g., influence diagrams, role maps, payoff matrices, heatmaps) to communicate the results clearly.

\subsection{Normal-Form Game Implementation}

The normal-form game process is organized into modular steps, each handled by a dedicated module or agent. First, a player-extraction module identifies the two players, their overall objective, the payoff unit (i.e., the measure in which outcomes are expressed, such as profit in USD, utility points, or years of freedom), and an expected value range from the user’s scenario description. Next, a strategy-generation module proposes up to ten strategies per player, ensuring concise names and one-line descriptions.
Based on the extracted strategies, an LLM-powered payoff assignment module constructs the payoff matrix by rating the outcomes of each strategy pair within the defined range. This matrix is then passed to a deterministic analyzer, which computes classical solution concepts: dominant strategies, Nash equilibria, and Pareto-optimal outcomes.  
Finally, the results are sent to an interpretation agent that summarizes the analysis in natural language, explaining the meaning of the equilibria, trade-offs, and dominant strategies. The agent concludes with a practical recommendation tailored to the user’s original scenario. Visual outputs such as payoff matrices or highlighted equilibria can be generated to further support understanding. 

\subsection{Sequential (Extensive-Form) Game Implementation}

The sequential-game pipeline mirrors the normal-form intake (players, objective, payoff unit, value range) and adds role conditioning: a role-assignment module creates concise “You are …” personas that become system prompts for player-specific LLM agents. Tree construction proceeds path-by-path from the root. Players alternate turns, at each node the active player’s agent proposes 1–3 concrete actions (or terminates if resolved), reasoning from its role rather than as a neutral narrator. A horizon policy ensures that paths remain meaningful and not too long: as the maximum depth is approached, the model avoids exploratory or redundant moves and instead favors termination. The maximum depth is a configurable parameter, allowing to balance between tree completeness and computational cost. All nodes are normalized into a consistent schema (decision or terminal), invalid outputs are corrected, and repeated actions along the same path are prevented. The result is a sequential game tree in which every path reflects alternating player choices grounded in their roles and limited by explicit, adjustable depth constraints.
After tree construction, a neutral evaluator (separate LLM agent) assigns payoffs at each leaf using the full path and given the objective, roles, and scenario as context. It returns a JSON payload with a brief rationale plus two numbers (one per player), enforcing the specified unit and value range. The payoff and reasoning are stored on the terminal node so every leaf remains auditable and quantitatively scored. Once payoffs are assigned, a deterministic analyzer computes Pareto-optimal and dominated paths, and derives a Subgame Perfect Equilibrium (SPE) via backward induction. The analyzer returns a compact summary (SPE path and payoffs, strategy profile per node, Pareto-optimal vs. dominated sets, and focus-player diagnostics). A final interpretation agent converts this summary into clear guidance: it explains the SPE’s strategic meaning, highlights Pareto trade-offs, clarifies why dominated paths are inferior, and names a single recommended decision consistent with the user’s objective and the tree’s outcomes. Optional visualization scripts can render the game tree with equilibrium highlights and path payoffs; these visuals are configurable and can be extended.

\subsection{Vester Sensitivity Analysis}

The Vester pipeline begins with an intake agent that transforms a natural-language scenario into a structured system definition. This includes: the system’s name and high-level description, its primary goal, explicit boundaries (what is inside vs.\ outside the system), functional subsystems, variables already mentioned by the user, and a stakeholder list (with at least five identifiable actors).  
For each stakeholder, a dedicated role-conditioned agent generates 3–5 candidate variables reflecting that stakeholder’s perspective for the system. Each variable is returned with a name, concise description, and optional category (economic, social, environmental, legal, or physical), ensuring that both direct and indirect influences are captured.  
The combined pool of variables is then passed to a deduplication and fit agent. This step merges near-duplicates under a representative name, consolidates descriptions, and records provenance via a merged\_from list. Importantly, user-mentioned variables are preserved explicitly. The outcome is a cleaned, system-consistent set of variables that aligns with the system’s boundaries and objectives, forming the foundation for subsequent steps of the sensitivity analysis. 
To construct the impact matrix, the system evaluates each ordered pair of variables \((A \rightarrow B)\) with two labels: an influence strength in \(\{0,1,2,3\}\) and an \emph{effect direction} in \(\{-1,+1\}\). The strength scale follows Vester’s rubric:  
\begin{itemize}
    \item 0 = no change at all,  
    \item 1 = a change in A brings about only a weak change in B,  
    \item 2 = A must change a lot to produce a comparable change in B,  
    \item 3 = even small changes in A cause large changes in B.  
\end{itemize}

The direction indicates whether an increase in A leads B to increase (\(+1\)) or decrease (\(-1\)).  
An LLM agent, primed to “mimic a stakeholder roundtable,” scores each pair by integrating perspectives from the system intake (system description, roles, goals, and variable definitions). Each evaluation is strictly directional and pair-specific, self-influence entries on the diagonal are forced to zero. To ensure consistency, the system sends concise “variable cards” (name, description, context) and requests only the integer score and sign. The result is a signed, weighted adjacency matrix that reflects interdependencies in stakeholder terms.
With the signed, weighted influence matrix, we compute for each variable its Active Sum (AS: row total) and Passive Sum (PS: column total). Using the dataset means \(\overline{AS}, \overline{PS}\) as dynamic thresholds, each variable is classified into the standard Vester roles: \textbf{Active} (AS>\(\overline{AS}\), PS\(\le\)\(\overline{PS}\)), \textbf{Critical} (AS>\(\overline{AS}\), PS>\(\overline{PS}\)), \textbf{Reactive} (AS\(\le\)\(\overline{AS}\), PS>\(\overline{PS}\)), and \textbf{Buffering} (AS\(\le\)\(\overline{AS}\), PS\(\le\)\(\overline{PS}\)). We also compute \(P=AS\cdot PS\) and \(Q=\frac{AS}{PS}\cdot 100\) (with \(Q=\infty\) if PS=0) for sensitivity dashboards. Additionally, we build a directed effect system and search for feedback loops using graph algorithms. Cycles are labeled positive (even number of negative edges) or negative (odd), and ranked by a Vester-oriented score that prioritizes stronger/shorter loops and increases with the share of \emph{Critical}/\emph{Active} variables and normalized \(P\). The analyzer returns: the ranked loop list, a compact “pretty” effect matrix (sign + strength) and an \emph{Top-$K$ critical loops} summary with variable frequencies for dashboarding. 
While the original Vester methodology also includes \emph{partial scenarios} and \emph{simulation}, our implementation does not cover these steps. We deliberately focused on the construction of the \textbf{impact matrix} and the subsequent \textbf{systemic role analysis}, which are widely regarded as the methodological core because they reveal the system’s key drivers, sensitive outcomes, and leverage points \cite{10.1007/978-3-319-07953-0_3}. Since our goal is to provide decision support directly from user prompts, the matrix and role analysis already supply the necessary structure for identifying interventions and reasoning about system dynamics. 
In the final step, an interpretation agent converts the full analysis into actionable guidance. It receives the user scenario, the cleaned context (deduplicated variables, stakeholders, subsystems, boundaries), the cross-impact matrix, sensitivity metrics per variable (AS, PS, \(P=AS\cdot PS\), \(Q=(AS/PS)\cdot 100\), role), the signed effect system, and the ranked feedback loops (including a Top-K “critical loops” set). The agent then briefly restates the system and scope, interprets sensitivity to identify drivers, sensitive outcomes, buffers, and critical leverage points, uses link signs to explain propagation (reinforcing vs.\ balancing), and proposes intervention priorities with expected side effects, a phased roadmap (quick wins vs.\ structural moves) and a single best lever recommendation. All claims are required to be grounded in the provided variables, roles, matrix links, and loop structures, ensuring an auditable, human-readable summary aligned with the computed artifacts.

\section{Evaluation}
We evaluate whether the proposed multi-agent LLM pipeline reproduces the structure and reasoning of a Vester-style sensitivity analysis on a real logistics case. As a human reference, we use a study from the IVL Swedish Environmental Research Institute (IVL) \emph{“A logistic analysis with the Sensitivity Model Prof. Vester”} \cite{wolf2012logistic}, which analyzes apparel flows from China to the Nordics via DK/SE entry points and a Swedish DC. The report proceeds through Vester’s nine steps, including variable elicitation across economic, environmental, social, organizational, and technical dimensions, a criteria filter, role charting, effect system, and partial scenarios; a second, later analysis adds company-internal factors. For like-for-like comparison, we treat the core transport-system model as the primary reference and the company/internal additions as a separate stratum.

\subsection{Methods}

\noindent\textbf{Prompting and runs.} We conduct 100 independent runs of the sensitivity analysis pipeline using a concise problem statement aligned with the IVL transport scope (no IVL variables disclosed). The pipeline configuration follows the \emph{Implementation} chapter (same modular steps and outputs), and all LLM agents across all steps use the same model with GPT-5.  

\noindent The exact user prompt was:  

\begin{quote}
\textit{For a real-world apparel transport flow from Shanghai to the Nordics (via DK/SE entry ports/airports and a Swedish regional DC to stores), how do the most consequential economic, environmental, social, organizational, and technical factors shape one another—and which of them emerge as primary levers, buffers/stabilizers, reactive indicators, and critical pressure points for system performance and sustainability?}
\end{quote}

\noindent\textbf{Semantic factor alignment.} To test whether the system surfaces conceptually equivalent factors, we compute sentence embeddings (all-mpnet-base-v2) for IVL factor names and for LLM factors (name+short label). For each IVL factor, the best LLM neighbor by cosine similarity is a match if $\geq 0.50$; otherwise it is unmatched. This threshold reflects typical behavior in modern embedding spaces, where unrelated items sit well below 0.5 while paraphrases or close conceptual equivalents exceed it, and is consistent with precedent in semantic similarity research \cite{10.5555/1631862.1631865}. Many-to-one mappings are allowed, since a single broader LLM factor can legitimately subsume several related human factors.

\noindent\textbf{What we evaluate.} 
(i) \emph{Factor alignment (match rate)} overall and by stratum (core vs.\ company/internal). 
(ii) \emph{Role consistency} for matched factors (agreement with IVL roles).

\textbf{Rubric-based judging.} In parallel, an LLM judge (GPT-5 Thinking) scores each run and a reconstructed packet of the IVL analysis under an eight-criterion rubric (max 100). The numbers in parentheses indicate the \emph{maximum points per criterion}: boundary clarity (5), stakeholder coverage (12), variable set completeness (12), causal structure (14), feedback loops (14), scenario design (14), temporal relevance (14), and actionability (15).

This evaluation adopts a rubric-based LLM judging approach, supported by recent evidence that advanced models can approximate human evaluation with high reliability: studies show GPT-4 aligns with human preference judgments in over 80\% of cases, comparable to inter-rater agreement \cite{zheng2023judging}. Leveraging this method provides a scalable and reproducible alternative to costly human rating campaigns while still grounding judgments in substantive criteria. In our setup, the goal is to grade the quality and internal coherence of each analysis (boundaries, factors, causal structure, loops, scenarios, etc.), not whether it copies IVL text. The same procedure is applied both to all LLM runs and to a reconstructed packet of the IVL baseline, enabling like-for-like comparison.  

\subsection{Results}

Across all 26 IVL factors, mean alignment was \textbf{55.5\%} (range \textbf{23.1–80.8\%}); \textbf{71\%} of runs reached $\geq 50\%$, \textbf{29\%} $\geq 65\%$, and \textbf{7\%} $\geq 75\%$.  
Restricting to the \emph{core transport-system} set (18 factors), mean alignment rose to \textbf{62.9\%} (range \textbf{27.8–94.4\%}; 5th–95th percentiles \textbf{38.6–83.3\%}). Here, \textbf{84\%} of runs reached $\geq 50\%$, \textbf{55\%} $\geq 65\%$, and \textbf{23\%} $\geq 75\%$.  
For the \emph{company/internal} additions (8 factors), mean alignment was \textbf{38.8\%} (range \textbf{0–75\%}; 5th–95th \textbf{12.5–62.5\%}); \textbf{30\%} of runs reached $\geq 50\%$.

At factor level, core items such as \emph{Fuel prices} (\textbf{99\%}), \emph{Competition in the transport market} (\textbf{95\%}), and \emph{Environmental demands of system actors} (\textbf{93\%}) were matched consistently, whereas breadth/structural measures (e.g., \emph{Total amount of goods}, \emph{Share of air freight}) and the internal set (e.g., \emph{Image of company}, \emph{Education and information}) matched less often—consistent with the transport-focused prompt boundary.

Role fidelity (over matched factors) averaged \textbf{56.6\%} overall (median \textbf{56.7\%}); aggregated per factor and averaged within strata, mean role match was \textbf{49\%} (core) and \textbf{41\%} (internal). High match rate did not guarantee role agreement for every factor; this sensitivity to small matrix differences parallels variation reported in participatory scenario work, where stakeholder mix can shape outcomes \cite{REED2013345}.

Table~\ref{tab:rubric-results} summarizes rubric outcomes across 100 runs versus the IVL baseline. LLM totals spanned \textbf{77–98} (mean \textbf{92.97}, SD \textbf{3.47}) against the human total of \textbf{93}. Boundary clarity (\textbf{4.99/5}), actionability (\textbf{14.94/15}), and temporal relevance (\textbf{13.5/14}) were consistently high; stakeholder coverage (\textbf{11.52/12}), variable completeness (\textbf{10.87/12}), and causal structure (\textbf{12.95/14}) clustered near the top. Greater spread appeared in feedback loops (\textbf{3–14}, mean \textbf{13.02}) and scenarios (\textbf{2–13}, mean \textbf{11.18}), aligning with IVL-level ranges.

\begin{table}[h]
\centering
\caption{Rubric scores from the LLM judge: 100 runs (summary) vs.\ IVL baseline.}
\label{tab:rubric-results}
\begin{tabular}{lcccccc}
\toprule
\textbf{Criterion} & \textbf{Max} & \textbf{Human} & \textbf{Mean (LLM)} & \textbf{SD} & \textbf{Min} & \textbf{Max} \\
\midrule
Boundary clarity     & 5  & 5  & 4.99 & 0.10 & 4  & 5 \\
Stakeholder coverage & 12 & 11 & 11.52 & 0.54 & 10 & 12 \\
Variable completeness& 12 & 11 & 10.87 & 0.48 & 9  & 12 \\
Causal structure     & 14 & 13 & 12.95 & 0.79 & 8  & 14 \\
Feedback loops       & 14 & 13 & 13.02 & 1.46 & 3  & 14 \\
Scenario design      & 14 & 13 & 11.18 & 1.41 & 2  & 13 \\
Temporal relevance   & 14 & 13 & 13.50 & 0.52 & 12 & 14 \\
Actionability        & 15 & 14 & 14.94 & 0.24 & 14 & 15 \\
\midrule
\textbf{Total}       & 100& 93 & 92.97 & 3.47 & 77 & 98 \\
\bottomrule
\end{tabular}
\end{table}

\subsection{Interpretation}

\noindent\textbf{What the alignment says.} The pipeline repeatedly surfaced the main external transport levers identified in the IVL report, indicating that, given only a scoped prompt, agents “thought” along similar systemic lines. Lower alignment for later company/internal factors reflects the evaluation scope: those factors emerged in a separate managerial workshop and are less likely to appear from an externally framed transport prompt.

\noindent\textbf{Roles are sensitive, as expected.} Moderate role agreement (\(\sim\)57\% over hits) is consistent with the inherent sensitivity of Vester’s method: even small changes in judged influences can shift variables across categories (active, reactive, buffering, critical). This is not a weakness of our approach but reflects the domain-specific nature of sensitivity analysis itself: role charts are highly dependent on influence ratings and stakeholder perspectives, and small changes in emphasis can legitimately produce different outcomes. In practice, such variability could also be expected in human workshops, where different expert groups or contextual assumptions may lead to alternative role classifications. While our pipeline reproduced this pattern, the result also highlights room for enhancement. Future work could explore diversifying models across steps to reduce variance and stabilize role agreement, while recognizing that some degree of variability is intrinsic to Vester’s design.

\textbf{Form, not imitation.} A key strength of our evaluation is that the analyses were judged not only on factor overlap but also on whether they were substantively well-formed Vester studies. The rubric-based evaluation by an LLM judge shows that the pipeline consistently produced analyses with the right structure and reasoning steps, rather than merely echoing surface features of the IVL report. Across 100 runs, total scores clustered tightly between \textbf{77} and \textbf{98} (mean \textbf{92.97}, SD \textbf{3.47}), nearly replicating the human baseline of \textbf{93} and in several cases even surpassing it. The narrow variance demonstrates that high-quality analyses were not isolated outputs but a stable pattern across runs.  
Taken together, these results show that the pipeline was not simply ``getting the right answers'' by chance, but was reliably following the \emph{form} of a Vester-style study. This demonstrates that an LLM pipeline can internalize the methodology itself: system boundaries, role logic, influence mapping, and feedback interpretation and apply it consistently across runs. That ability to replicate structure and reasoning rather than text is a central strength of our approach, and the reason why the pipeline can be considered not just an imitation of human output, but a genuine operationalization of the method.

\noindent\textbf{In summary}, the evaluation shows that the LLM system can approximate the structure and reasoning of a participatory Vester analysis in a complex logistics domain. Each run constitutes an individual analysis and is therefore not identically reproducible, just as human workshops are not: changing experts, emphasis, or minor influence judgments yields legitimate variation in factors and roles, consistent with evidence that stakeholder composition can materially alter outcomes \cite{REED2013345}. However, the system consistently identified the key external transport-system levers, assigned roles that overlapped substantially with the human workshop, and achieved high rubric scores from an impartial LLM judge (\textbf{77–98}, mean \textbf{92.97}, SD \textbf{3.47}), matching the human based baseline of \textbf{93}. Alignment gaps for internally oriented factors and partial divergences in role fidelity highlight the dependence on framing and available knowledge, but these are precisely the aspects where expert-system augmentation could strengthen performance. Overall, the results suggest that multi-agent LLM pipelines are not only able to “think” along the same systemic lines as human experts but, with access to domain-specific context, may evolve into robust tools for participatory-style sensitivity analysis across specialized fields.

\section{Conclusion and Outlook}
This paper implemented and demonstrated three explainable analysis processes within a modular LLM-agent system: Vester’s sensitivity analysis, normal-form game analysis, and sequential (extensive-form) game analysis. Each process externalizes intermediate artifacts (influence matrices and role charts; payoff tables; decision trees with backward-induction summaries), thereby replacing opaque one-shot answers with auditable reasoning steps. 

Our evaluation shows that the approach can reproduce the form and reasoning of established human methods rather than yielding coincidental overlaps. On a logistics case study, we executed the Vester pipeline 100 independent times and compared against a human baseline. Mean factor alignment across all 26 IVL factors was 55.5\% (core transport subset: 62.9\%; company/internal subset: 38.8\%). Role fidelity over matched factors averaged 56.6\%. A rubric-based LLM judge rated structure and methodological quality tightly around the human baseline (LLM mean 92.97/100 vs.\ human 93/100), indicating that outputs were consistently well-formed Vester studies.

Revisiting the opening problem from our introduction, the Canada housing crisis, we applied our framework to this scenario. We provided the system with a problem statement capturing chronic undersupply, rising costs, permitting delays, financing risks, and their socio-economic consequences. When processed through our Vester analysis pipeline, the salient levers and risks closely aligned with those emphasized in the federal initiative \emph{Build Canada Homes (BCH)}—a new agency launched on September 14, 2025 to accelerate affordable and middle-class housing.\footnote{Government of Canada: ``Prime Minister Carney launches Build Canada Homes,'' September 14, 2025. Available at: \url{https://www.pm.gc.ca/en/news/news-releases/2025/09/14/prime-minister-carney-launches-build-canada-homes}}

Concretely, our analysis foregrounded public land availability, approval and permitting speed,  financing risk under volatile interest rates, and the capacity of modular construction and skilled labour. These map directly to BCH’s pillars: leveraging federal public lands, streamlining approval processes, mobilizing private and non-profit developers with financial instruments, and scaling modern construction methods. Both perspectives converge on the same critical uncertainties, construction cost inflation, interest rate trajectories, regulatory and intergovernmental coordination, supply chain and labour bottlenecks, land servicing requirements, and the operational definition of ``affordable.''

It is important to note that the LLM had no prior knowledge of these specific policy measures: neither the government program itself nor its details were present in the provided prompt, and they were outside the model’s training knowledge, which currently has a cutoff of September 30, 2024.\footnote{OpenAI Documentation: ``GPT-5 models,'' knowledge cutoff September 30, 2024. Available at: \url{https://platform.openai.com/docs/models/gpt-5}} This correspondence therefore illustrates that the LLM based sensitivity analysis surfaced actionable levers consistent with contemporary policy design, rather than offering vague, black  box prescriptions.

Taken together, this marks a step toward XAI via implementing traceable frameworks: multi-agent LLM pipelines can reproduce known methods with traceable steps, deliver human-aligned and inspectable decision support.

For future work, our results suggest that small changes in influence ratings can shift systemic roles, highlighting the sensitivity of these frameworks to initial judgments. We therefore propose an \emph{expert layer} that can be flexibly integrated into each process of the pipeline. Due to the modular design of our implementation, such an expert layer could be selectively activated at steps where external validation or additional context is most valuable. 

The expert layer is not limited to human input. It could draw from multiple complementary sources, such as real-time web search to inject up-to-date information, domain-specific or fine-tuned models that encode sector knowledge, structured company data or transcripts from internal meetings in organizational settings, or curated expert feedback at critical decision points. 
By enriching the agent processes with targeted knowledge at high leverage steps, the expert layer has the potential to improve both the robustness and fidelity of the analysis, while maintaining overall autonomy and scalability of the system.
Our evaluation used a single model configuration across steps. Future experiments could diversify models per step (e.g., retrieval-heavy steps vs.\ numeric-judgment steps), test reasoning-focused variants, and apply cross-model self-consistency (vote/aggregate) to stabilize outcomes.

% ---- Bibliography ----
% LNCS: use splncs04 (this shortens long author lists to “et al.”)
\bibliographystyle{splncs04}
\bibliography{reference2}

\end{document}